# Fake News Detection: Experiments and Approaches beyond Linguistic Features

**Shaily Bhatt[1], Sakshi Kalra[2], Naman Goenka[3], Yashvardhan Sharma[4]**

[1,2,3,4] Web Intelligence and Social Computing Lab,
Department of Computer Science and Information Systems,
Birla Institute of Technology and Science, Pilani,
Pilani – 333031, Rajasthan, India

[1] f20170040@pilani.bits-pilani.ac.in, +91-9119225096

[2] p20180437@pilani.bits-pilani.ac.in, +91-7888929069

[3] f20180398@pilani.bits-pilani.ac.in, +91-8824145677

[4] yash@pilani.bits-pilani.ac.in, +91-9660048176

## Abstract

*Easier access to the internet and social media has made disseminating information through online sources very easy. Sources like Facebook, Twitter, online news sites and personal blogs of self-proclaimed journalists have become significant players in providing news content. The sheer amount of information and the speed at which it is generated online makes it practically beyond the scope of human verification. There is, hence, a pressing need to develop technologies that can assist humans with automatic fact-checking and reliable identification of fake news. This paper summarizes the multiple approaches that were undertaken and the experiments that were carried out for the task. Credibility information and metadata associated with the news article have been used for improved results. The experiments also show how modelling justification or evidence can lead to improved results. Additionally, the use of visual features in addition to linguistic features is demonstrated. A detailed comparison of the results showing that our models perform significantly well when compared to robust baselines as well as state-of-the-art models is presented.*

## Keywords

# 1. Introduction

Social media for news consumption is a double-edged sword. The minimal effort, simple access, and quick dispersal of data on the internet and social media are increasingly encouraging people to switch from traditional sources of news to online ones. Sources like Facebook, Twitter, online news sites, other social media platforms and personal blogs of self-proclaimed journalists have become significant players in providing news content. The sheer amount of information and the speed at which it is generated and propagated online makes it practically beyond the scope of human verification. There is, hence, a pressing need to develop technologies that can assist humans with automatic fact-checking and reliable identification of fake news.

The term "fake news" is not new. Journalists often define fake news as referring to "viral posts based on fictitious accounts made to look like news reports" [7]. A recent study defined fake news "to be news articles that are intentionally and verifiably false and could mislead readers" [3]. While what categorizes as fake news is an open debate and has a broad spectrum of social, psychological, and factual perspectives attached to it. However, limiting to a simplified definition of fake news as "content that has been intentionally created to mislead the readers" will suffice for the scope of this paper [3].

Here, an empirical study to develop fake news detection models by leveraging the latest breakthroughs of deep learning, more specifically, neural networks for processing language and visual data is presented. The main contribution of this paper is to substantiate the three hypotheses: First, features other than those of language can significantly improve performance of fake news detection models. Metadata and information about the author/speaker have been employed to create a credibility index which together with linguistic features leads to significant improvement in results. Second, modelling evidence or justification, along with the supervised labels of the articles, improves the model performance significantly. Finally, visual features can benefit the task performance. For substantiating this experiments on visual features that can be exploited for creating multimodal fake news detection models are presented. Experiments have been carried out on standard datasets which are well accepted within the research communities. These include the LIAR, LIAR-Plus and FakeNewsNet. The techniques used range from traditional machine learning based approach for establishing a baseline to more advanced and complex neural networks. Throughout the paper, results are compared with robust baselines and state-of-the-art models.

# 2. Related Work

## 2.1. Definition of Fake News

What categorises as fake news is open to debate on a variety of levels such as psychological, social and contextual foundations [3]. Authenticity and intent are two critical features in Fake News [3], and thus a narrow definition of fake news is "intentionally written misleading content". A similar definition of Fake News is "content whose veracity is compromised by intentional deception" [7]. From the perspective of fact-checking, fake news can be defined as "news that is verifiably false". A broader definition would include satire, parodies, unverifiable claims and unintentional rumours as fake news. What categorises as fake news is an open and evolving debate that has a broad spectrum of social, cultural and psychological aspects tied to it. The most common feature accepted by all definitions is "the intentional spread of verifiably false information". For the scope of this study, the narrow definition of fake news defined as "content created with the intention of misleading readers" will be followed.

## 2.2. Challenges in Automatic Fake News Detection

A comprehensive analysis of the challenges faced in automatic fake news detection is described in [13]. The various challenges faced are:

First, the involvement of multiple players in the news ecosystem increases the difficulty of building computational, technological and business strategies that can cope with the dynamic and quality information.

Second, malicious or adversarial intent is tough to detect. Thus, it is difficult to segregate false information that is spread with the intent of misleading readers from the news that contain false information due to honest mistakes.

Third, the lack of public awareness and vulnerability of the audience plays a crucial role in the dissemination of false information.

Fourth, social and cultural differences play a role in psychological and contextual interpretation. There may be differences in perspectives which make it challenging to categorize news articles as fake or real. Such news also attacks vulnerable emotions of the audience, hence determining veracity from the style of news content may prove to be unreliable.

Furthermore, the matter is complicated by the dynamic nature of the process of propagation of fake news through the internet and social media. False information spreads at tremendous rates and changes rapidly as it passes from one user to another.

Finally, a significant challenge is posed by fast-paced developments in the world. Knowledge-based systems need to retrieve information on newly emerging facts continuously. This causes static models to suffer from what is known as 'concept drift' in machine learning models which means that the data on which the model was trained becomes obsolete.

## 2.3. Existing Datasets

There are several standard datasets publicly available for the research community [4, 12, 13, 15]. The most used datasets include BuzzFeedNews, BS Detector, PHEME, CREDBANK, BUZZFACE, FacebookHoax, LIAR and FakeNewsNet.

The major challenges faced in the creation of datasets from these techniques are: First, crowd-sourced datasets have a degree of doubt associated with the ground truth label itself. This makes models built of such datasets unreliable. Second, there is no algorithm to label websites generating news content as malicious or authentic. The probability of both false positives and false negatives is significant, again making datasets obtained from such websites unreliable. Furthermore, fact-checking websites often focus on specific topics like politifact.com is for political news only. Thus, it is not possible to obtain comprehensive datasets from such websites. Finally, expert fact-checking and human annotation is extremely time-consuming as well as costly.

In this paper the LIAR dataset [5] is used for experimentation on linguistic features with source/author credibility and metadata. The LIAR-Plus [6] which is built on top of the LIAR dataset and contains additional justification or evidence for the label associated with the news is also used. For experiments involving visual features the FakeNewsNet dataset [10] which contains both news pieces and images, among other features is utilised.

## 2.4. Classification Methods

Recent literature considers Fake News Detection as a classification problem: their goal is to provide labels fake or real to a particular news piece. In many of the cases, authors have used machine learning, both supervised and unsupervised, and deep learning methods. Other scholars have applied data mining techniques, time series analysis, and have utilised external resources (e.g., knowledge bases) to assess their credibility. Features used include those from linguistic analysis, semantic and contextual understanding of language, metadata, multimodal data, network analysis, among others.

Authors in [19] report a traditional machine learning based technique. K-means is used for feature selection, and a supervised learning based technique, Support Vector Machine (SVM) has been used to classify the fake news from the corpus.

The paper [16] addresses the problem of labelled benchmarked datasets by applying a two-path semi-supervised technique. One of the paths is supervised, and the other is unsupervised. For the extraction of the features, a shared CNN has been used. Both the paths are jointly optimised to complete semi-supervised learning.

The authors of [18] have proposed a model called FNDnet, which leverages a deep convolutional model for classification. The model achieves the highest accuracy of 98.36 comparable to the state of art methods by evaluation on the Kaggle Fake News dataset. The limitation, however, in this case, is that the model was not tested with other benchmark datasets which are commonly used and accepted by the research community.

The paper [17] addresses the problem by applying a capsule neural network which is has been previously used in the computer vision tasks and is now receiving attention for use in language tasks. Different embedding models for news items of different lengths have been used, and distinct levels of n-grams have been used for the feature extraction. The model has been tested on LIAR, and ISOT datasets and performance better than state-of-the-art is reported in the paper. Comparison of our model performance to this model is presented in later sections.

The authors of [20] propose Fakedetector, a novel deep diffusive neural network and perform experiments on a dataset obtained from politifact.com. They obtain the best accuracy in comparison to a number of competitive methods that use textual information for prediction. The dataset used in their experiments has been obtained from the same website from which LIAR and LIAR-Plus have been created. Hence, comparisons of our models to the Fakedetector model described in this paper is given in later sections.

In the paper [19] author reported a novice multimodal architecture by considering both the text and image features and model is evaluated on the self-generated dataset named r/Fakeddit, which is collected from Reddit. Pre-trained InferSent and BERT have been used for the text feature extraction and VGG16, ResNet 50 and EffificentNet have been used for image feature extraction. We plan to test our multimodal model on this dataset in the future.

In [14], authors develop a novel network, a Multimodal Variational Autoencoder (MVAE) to learn features from text and images jointly. The network learns probabilistic latent variable models and couples it with a binary fake news classifier. The model is tested on data from datasets obtained from Twitter and Weibo and reports state-of-the-art results.

Authors in [21] have built a multimodal architecture, Similarity-Aware Multimodal Fake News Detection model, SAFE, that considers the relationship or similarities between the textual and visual information in the news articles. First, a neural network is used for the text and image feature extraction. Secondly, the relationship among the extracted features across different modalities is investigated and based on similarities and mismatches, and news article is classified. This model has been tested on the FakeNewsNet dataset and outperforms baselines and competitive models to give the best performance in all cases. Comparison of our multimodal results to those obtained in this paper is presented in later sections.

## 2.5. Evaluation Metrics

By formulating this as a classification problem, standard metrics defined are: Precision, Recall, Accuracy and F1 score. In binary classification, fake news is taken to be a positive class. Thus, True Positive (TP) is when predicted fake news piece is actually fake, True Negative (TN) is when predicted true news is actually true, False Positive (FP) is when a predicted fake news is actually true and False Negative (FN) is when predicted true news is actually fake. Thus, the formulas for calculating the above four metrics are:

$$Precision = \frac{|TP|}{|TP| + |FP|} \qquad Recall = \frac{|TP|}{|TP| + |FN|}$$

$$Accuracy = \frac{|TP| + |TN|}{|TP| + |TN| + |FP| + |FN|} \qquad F1\ Score = \frac{2 \times Precision \times Recall}{Precision + Recall}$$

Different perspectives for a classifier can be drawn from these evaluation metrics. Specifically, accuracy measures how much actual fake news and predicted fake news are similar in terms of the features. Precision addresses the important problem of identifying which news is fake by measuring the fraction of all detected fake news that are annotated as fake news. Often, fake news detection datasets skewed, allowing fewer positive predictions to result in a high precision. Recall is the fraction of actual fake articles predicted as fake. This is used to measure the sensitivity of the classifier. F1 provides an overall performance measure by combining precision and recall. Better performance of the classifier is indicated by a high value of each of the evaluation metrics.

# 3. Description of Datasets Used

For the experiments of this paper, the datasets used are: LIAR[5], LIAR-Plus[6] and FakeNewsNet[10]. The focus is on the news content and auxiliary features of the dataset. Contextual and linguistic features from both datasets are explored, extracted and modeled. Visual features are utilised from FakeNewsNet.

The LIAR dataset contains features including statement (or claim), label (six-classes) subject, speaker and auxiliary information about the speaker such as job, political affiliation, state, context/venue of the claim and credit counts. The LIAR-Plus [5] builds on this by adding a justification column where evidence is provided for why a particular claim is labelled into a particular category. This information is referred to as as evidence or justification throughout the paper. Both these datasets have a six-grained labelling with classes as – True, Mostly True, Half True, Barely True, False and Pants-on-fire.

The FakeNewsNet [10] repository has a multitude of features including news content features, network features and spatio-temporal information. In this paper the is focus on the news content by exploiting visual-based features along with linguistic and contextual features.

# 4. Experiments and Results

Several techniques of increasing complexity and features on the above-mentioned datasets are explored in our experiments. First, four models on the LIAR and LIAR-Plus dataset that take into consideration only linguistic features are proposed. These are - regression model, Siamese network with BERT base [8], sequence model and an enhanced sequence model. Then, experiment with two models on the FakeNewsNet dataset are presented, one of which uses only contextual linguistic features (Sequence model) and another that additionally uses visual features (CNN model). This provides us with a comparison of whether visual features are useful in distinguishing fake news. The specifics of techniques and results obtained for each of the techniques are described as follows.

## 4.1. Regression Model

To begin with one of the most preliminary machine learning technique: regression is used. The purpose of this experiment is to establish a baseline for automatic fake news detection. GLoVe embeddings [1] are used to encode the text. This model is tested on the LIAR-Plus Dataset. Standard pre-processing steps like removal of stop words, neglecting casing, substituting missing values with average and ignoring words not present in GLoVe were applied. Note – these pre-processing steps are applicable to all models described in future sections also.

The regression models used are Linear Regression (LR), Logistic Regression – one vs rest (LoR) and Ordinal Logistic Regression (OLR). The results obtained on 5-fold cross validation gave *highest mean accuracy of 31.57% on six-way classification*. The model is adapted to binary classification. The classes – pants-on-fire, false and barely true are categorized into one class – FALSE and half true, mostly true and true are categorized into one class – TRUE. The *highest mean accuracy on 5-fold cross validation is 65%.* The results are shown in Table I.

**Table I**
**Results obtained on LIAR-Plus using Regression models**

| Model | Six-way classification | | Binary classification | |
|---|---|---|---|---|
| | Mean Accuracy | Variance | Mean Accuracy | Variance |
| LR | 0.2287 | 4.092e-05 | 0.6500 | 6.546e-06 |
| LoR | 0.3157 | 4.396e-05 | 0.6321 | 1.050e-05 |
| OLR | 0.2384 | 3.76e-05 | 0.6500 | 6.266e-06 |

## 4.2. Siamese Network with BERT

An artificial neural network that uses the same weights while working on two different inputs to produce a comparable output is known as a Siamese Neural network. The following Siamese models with BERT in the base architecture were used on the LIAR-Plus dataset. One Branch, Two branch and Triple Branch Siamese network were utilised. Best results were obtained with the Triple branch network. The model architectures shown in Figure 1 are described below:

1. **Single Branch**: The input sequence is first passed through a pre-trained BERT model. The BERT architecture is fine-tuned by passing the tensor from BERT through a linear fully-connected layer (FC). This gives a binary output for fake or true labels. Here, no metadata or justification is used for training; only a single branch with the news statement is used. ***Testing accuracy of 60% is obtained on binary classification***.

2. **Two Branch**: News statement and justification were used to create the two branches of this network. These are passed through a linear FC layer after concatenation. Both these branches share weights. This architecture makes use of the 'justification' along with statement giving better result than the single branch. ***Binary classification accuracy of 65.4% and six-way classification accuracy of 23.6% was obtained using this method***. As evident, this is an improvement from the single-branch model substantiating the hypothesis that using justifications can give better results.

3. **Triple Branch**: An additional branch is added in this approach. This branch takes as input the additional available metadata such as speaker, source, affiliation etc. The authenticity of the publisher is considered using the feature *"Credit Score"*(CS) as defined in equation (1). The CS was added to the concatenation of output from the three branches. The length of the input sequence of each branch is modified to be equal the average input length in that branch.

   The six-way classification accuracy improved by a huge margin. ***37.4% and 77.2% was the highest accuracy obtained for six-way and binary classification, respectively.*** The binary classification accuracy is 7.2% higher than the accuracy obtained by authors.

As evident, the accuracy of two branch network is higher than one branch substantiating the claim that justification modelling can lead to improved accuracies. Further the triple-branch accuracy is even better as both metadata and justification are used. Once again, this proves the hypotheses that features beyond linguistic ones can help improve model performance. There is a scope for further improvement by finding better methods to integrate metadata and credit score and further fine-tuning the model. These results are further improved in the next experiment with sequence models and enhanced sequence model.

**Figure 1**
**Architecture of BERT Based Siamese Network**

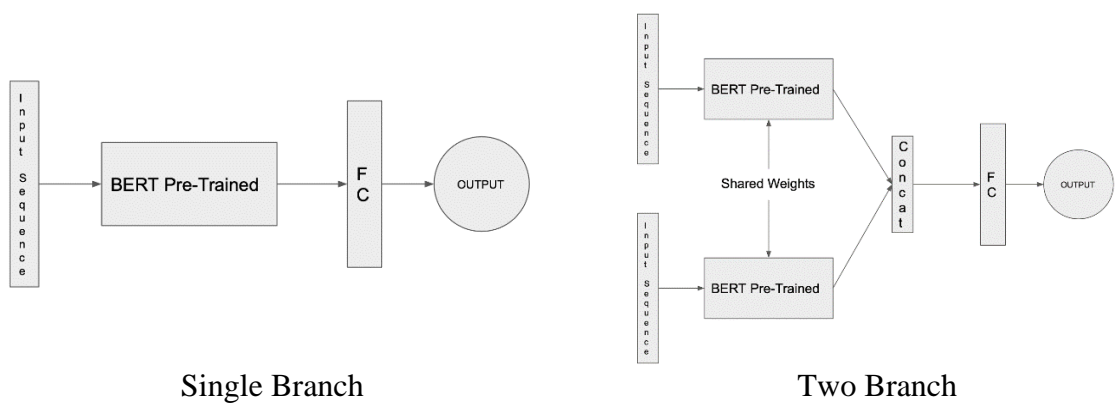

Single Branch                                        Two Branch

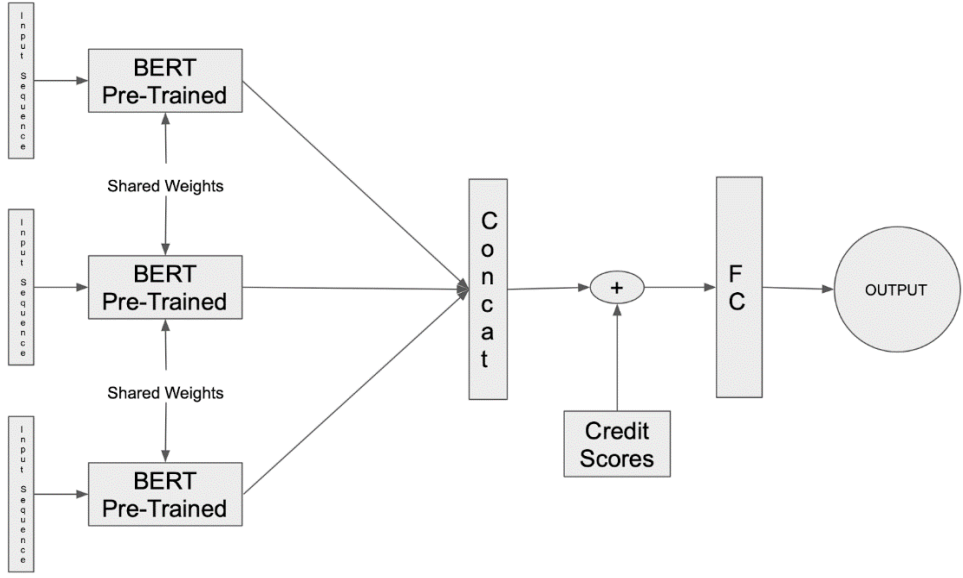

Triple Branch

## Definition of Credit Score (CS)

The scalar – Credit Score (CS) - is indicative of the credibility of the author, calculated as counts of false news propagated by the author in the past. A weighted aggregate of the six-grained counts provided for every author followed by tanh activation is used to calculate CS. The weights are taken as hyper-parameters and not tuned by the model. The scalar is defined as in equation 1.

$$CS = \tanh\left[w * \left(\frac{0.2 * MTC + 0.5 * HTC + 0.75 * BTC + 0.9 * FC + 1 * PFC}{MTC + HTC + BTC + FC + PFC}\right) + b\right] \quad (1)$$

Here, MTC refers to mostly true count for the speaker, HTC refers to half-true counts, BTC refers to barely true counts, FC refers to false counts and PFC refers to pants-on-fire counts. The credit score is passed to a 1-neuron dense layer to learn the relative importance of credit score in determining the final claim and w and b are weight and bias learned during training.

The scalar is biased towards authors with more false counts due to progressively higher value of weights from mostly-true to pants-on-fire counts. The rationale behind choosing such a weighting scheme is that the knowledge about an author making a false statement is intuitively more critical for judging the credibility of his/her statements. The credit score is useful in distinguishing fake and real news by creating a relative difference in the activation outputs because more the credit score, less reliable is the person making a claim.

## 4.3. Sequence Model

Sequence Models have been used in several Natural Language Processing tasks. Two models trained here are: one without using the justifications as in LIAR and another using the justification from LIAR-Plus. The data was pre-processed with standard techniques like removing stop words, neglecting casing, substituting missing values with average etc. GLoVe vector embeddings [1] of dimension 100 were used to input the statements and justifications, and the rest of the data was fed to a feed-forward neural network.

The architectures of these models have been shown in Figure 2. In both the architectures, the first input branch is for the encoded statement and the second input branch is for speaker-related metadata. The third input branch in the model with justification corresponds to the encoded justification from LIAR-Plus. In the model architecture shown in the figure: LSTM nodes refer to a standard LSTM layer with 128 cells. Dropout nodes refer to regularisation dropout with probability 0.15 in the statement branch and 0.2 in the justification branch to prevent overfitting. Dense nodes refer to fully connected dense feed-forward layer with 32 units in the statement and justification branch and 64 units in the metadata branch followed by Relu activation. Concatenate node is a concatenation layer to combine outputs of each branches. The final dense node after concatenation is a feed-forward layer with softmax activation for output.

The binary classification model was trained for 120 epochs with a batch size of 512. The six-grained classification model is trained for 40 epochs. The data was distributed into: 16000 Training rows, 4000 Validation rows, 1744 Testing rows. The results are shown in table II.

**Figure 2**
**Architecture of Sequence Models**

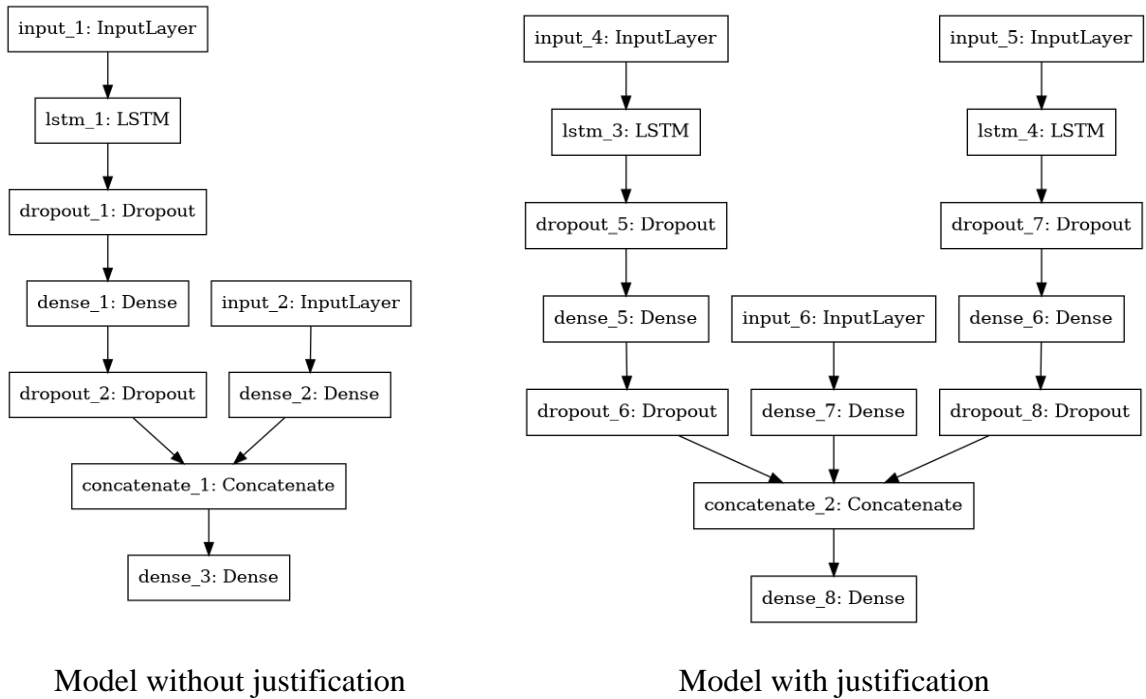

Model without justification                    Model with justification

**Table II**
**Results obtained by Sequence Model**

| Classification | Dataset | Justification | Accuracy | |
|---|---|---|---|---|
| | | | **Training** | **Testing** |
| Binary | LIAR | No | 0.8192 | 0.7862 |
| | LIAR-Plus | Yes | 0.8559 | 0.8205 |
| Six-way | LIAR-Plus | Yes | 0.5439 | 0.5015 |

As expected, sequence model shows significant improvements on both binary and six-grained classification in LIAR as well as LIAR-Plus datasets. Once again as expected accuracy performance with justification is better in both binary and six-way classification.

## 4.4. Enhanced Sequence Model

The sequence model is enhanced by introducing an additional branch with the "Credit Score" (CS) as defined in equation 1. The model architecture is shown in Figure 3. The four branches have inputs as statement (S branch), metadata (M branch), justification (J branch) and Credit Score (C branch). The sizes of input layers are as mentioned in the figure. The LSTM node is a standard LSTM layer with 128 cells in the S and J branch. The dropout node is regularisation dropout to prevent overfitting in the S and J branch with a dropout probability of 0.15 and 0.21 respectively. The dense layer is a feed-forward fully connected layer with 32 units each in the S and J branch, 64 units in the M branch and 1 unit in the C branch. This is followed by a Relu activation in the first three branches and tanh activation with the credit score as defined in the definition of credit score. The concatenate node concatenates the results of the S, M and J branch which is then passed to the Add node which adds it to the result from the C branch. Finally, a single dense layer with sigmoid in case of binary classification and softmax in case of 6 classes is used to give the final label output.

Categorical cross-entropy loss and ADAM optimiser were used for training this model for multi-class fine-grained classification. Binary cross-entropy loss and ADAM optimiser were used for training this model for binary classification. Early-Stopping in Keras callback with the patience of 15 epochs and validation loss being monitored quantity was used to prevent overfitting. For binary classification, model was trained for 500 epochs and for multi-classification, the model was stopped early on 230th epoch. ADAM learning rate was tuned after grid search to 0.001 for both classification tasks.

All results were verified using 5-fold stratified cross-validation in both classification tasks. Batch size was tuned after grid search to 256. Table III outlines our best results achieved. Figure 4 shows model loss for binary and six-way classification for the best model.

This performance is a significantly better than the scores reported in the papers [5, 6] where the authors first introduced the LIAR and LIAR-Plus dataset. These results clearly show that using a weighted aggregate of credit score can give better performance. Finally, comparison of our results to the performance of Fakedetector [20] which reports state-of-the-art performance (to the best of our knowledge) after comparison to various baselines and competitive models is described here. The authors of this paper use a dataset with ~14k examples (slightly less than the ~16k examples in LIAR-Plus) which have also been obtained from PolitiFact.com along with justification and other metadata. This is a fair comparison due to strong similarity in dataset size, features available as well as dataset sources.

Fakedetector obtains a maximum accuracy of 0.64 on binary classification and 0.29 on six-way classification. ***Our scores using the Triple Branch Siamese network as well Enhanced Sequence Model exceed these results.*** This improvement can be attributed to the effectiveness of the newly introduced credibility index – Credit Score (equation 1) which is not used in the case of Fakedetector.

**Table III**
**Results of Enhanced Sequence Model**

| Classification | Accuracy | | Precision | Recall | F1 score |
|---|---|---|---|---|---|
| | Training | Testing | Testing | Testing | Testing |
| Binary | 0.8403 | 0.8297 | 0.734 | 0.712 | 0.722 |
| Six-way | 05370 | 0.5272 | 0.43* | 0.42* | 0.42 |

*Refers to macro average for all six classes

**Figure 3**
**Architecture of Enhanced Sequence Model with details of layers and hyperparameters**

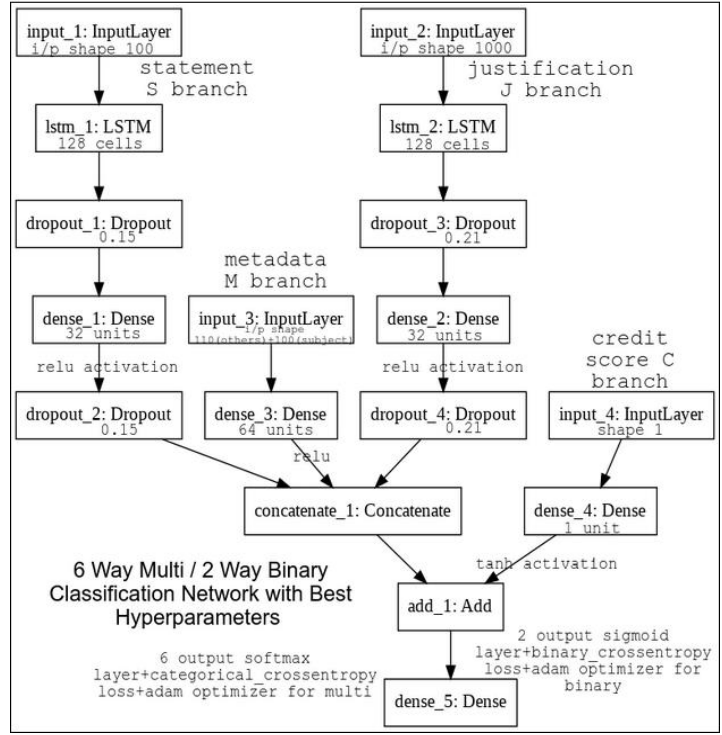

## 4.5. Sequence Model on FakeNewsNet

The FakeNewsNet dataset contains a variety of features including linguistic and visual features. First, experiments with a sequence model similar to the one used in the previous sections are presented in order to establish a baseline score for comparison of the multimodal model.

***F1 score of 98.74% is obtained on training and 93.71% is obtained on validation for binary classification.*** The scores of this sequence model are not compared with those of the previous one because of the different datasets used in both cases. The scores of models trained on FakeNewsNet are better owing to the larger number of examples in the dataset. The dataset also does not have justification or six-grained labels and hence those comparisons are not made. The purpose of this model is solely to establish a baseline against which we the performance of multimodal model will be compared. The model loss is shown in Figure 4.

## 4.6. Convolutional model for Linguistic and Visual features

A CNN is trained on the features of text and images simultaneously [11]. Multiple convolutions are employed to capture the hidden features of text and images. Features are classified as latent (hidden) or explicit. There are two parallel CNNs to extract features from text and images and text, respectively. The latent and explicit features are then projected on the same feature space. These representations are then fused to give output.

The model loss is depicted in Figure 5. **_F1 score of 99.2% is obtained in training, and 96.3% is obtained in validation._** It is clearly evident from these results that using the visual features in addition to linguistic features in fake news detection can lead to performance enhancement. This substantiates the assumption that images that are often associated with news posts on social media can be an important indication for the veracity of the news item.

Finally, a comparison of our results to state-of-the-art performance reported in literature is described as follows. The MVAE, a multimodal model proposed in [14] reports an F1 score of 73% on a multimodal dataset from Twitter and 83.7% on a multimodal dataset from Weibo. Our model gives a better F1 performance. However, this difference maybe partly due to the difference in datasets. The authors in [21] propose a multimodal model, SAFE which they rigorously test on the FakeNewsNet dataset. They compare their models to multiple baselines as well as competitive models and report state-of-the-art performance (to our best knowledge) with an F1 score average of 0.8955. Our model improves this performance by 6 points using the parallel CNN method described in this section.

**Figure 4**
**Loss of Sequence Model on FakeNewsNet**

**Figure 5**
**Loss of CNN Model on FakeNewsNet**

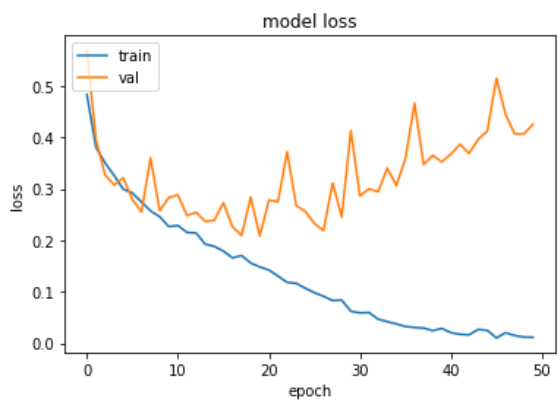
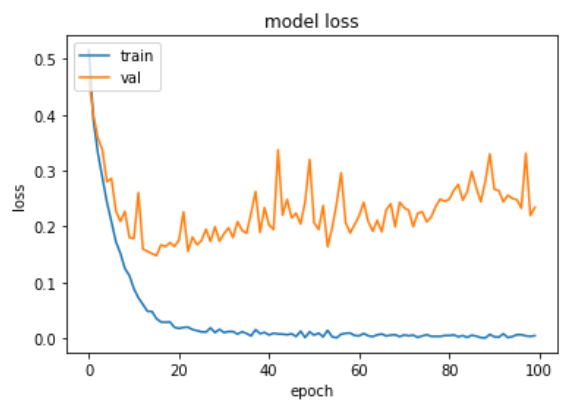

## 5. Conclusion

In this paper, an empirical study of multiple models of varied complexities for fake news detection is presented. We attempt to go beyond using just linguistic features to improve model-performance on the task. The hypotheses proved using experiments are: First, credibility index of source/speaker and metadata associated with news articles can play a crucial role in improving model performance. Second, modelling evidence or justification with the news claim significantly improves the model performance. Finally, multi-modal models that exploit visual features from images associated with news articles can perform better than models that utilise only linguistic and contextual features.

Experiments are carried out using the LIAR, LIAR-Plus and FakeNewsNet datasets and comparisons of our results with baselines and state-of-the-art models is presented. Our models gave results comparable and better than state-of-the-art models.

Best accuracy is obtained on binary classification and six-way classification on the LIAR-Plus dataset using an enhanced LSTM-based sequence model which uses linguistic features, credit scores, metadata and justification. An improvement in performance on the FakeNewsNet dataset is obtained using a multi-modal model as compared to the model that uses only linguistic features.

Finally, the findings in the case of the multi-modal model are particularly encouraging. In the future, we plan to explore the further integration of features from visual data for building better fake news detection systems.

# References

[1]  Pennington, Jeffrey, Richard Socher, and Christopher D. Manning. "Glove: Global vectors for word representation." In Proceedings of the 2014 conference on empirical methods in natural language processing (EMNLP), pp. 1532-1543. 2014.

[2]  Conroy, Nadia K., Victoria L. Rubin, and Yimin Chen. "Automatic deception detection: Methods for finding fake news." Proceedings of the Association for Information Science and Technology 52, no. 1 (2015): 1-4.

[3]  Allcott, H., & Gentzkow, M.: Social media and fake news in the 2016 election. Journal of economic perspectives, 31(2), (2017) 211-36.

[4]  Shu, Kai, Amy Sliva, Suhang Wang, Jiliang Tang, and Huan Liu. "Fake news detection on social media: A data mining perspective." ACM SIGKDD explorations newsletter 19, no. 1 (2017): 22-36.

[5]  Wang, William Yang. ""Liar, Liar Pants on Fire": A New Benchmark Dataset for Fake News Detection." In Proceedings of the 55th Annual Meeting of the Association for Computational Linguistics (Volume 2: Short Papers), pp. 422-426. 2017.

[6]  Alhindi, Tariq, Savvas Petridis, and Smaranda Muresan. "Where is your Evidence: Improving Fact-checking by Justification Modeling." In Proceedings of the First Workshop on Fact Extraction and VERification (FEVER), pp. 85-90. 2018.

[7]  Tandoc Jr, E. C., Lim, Z. W., & Ling, R.: Defining "fake news" A typology of scholarly definitions, Digital journalism, 6(2), 2018, 137-153.

[8]  Devlin, Jacob, Ming-Wei Chang, Kenton Lee, and Kristina Toutanova. "Bert: Pre-training of deep bidirectional transformers for language understanding." arXiv preprint arXiv:1810.04805 (2018).

[9]  Pérez-Rosas, Verónica, Bennett Kleinberg, Alexandra Lefevre, and Rada Mihalcea. "Automatic Detection of Fake News." In Proceedings of the 27th International Conference on Computational Linguistics, pp. 3391-3401. 2018.

[10]  Shu, Kai, Deepak Mahudeswaran, Suhang Wang, Dongwon Lee, and Huan Liu. "Fakenewsnet: A data repository with news content, social context and dynamic information for studying fake news on social media." arXiv preprint arXiv:1809.01286 8 (2018).

[11]  Yang, Yang, Lei Zheng, Jiawei Zhang, Qingcai Cui, Zhoujun Li, and Philip S. Yu. "TI-CNN: Convolutional neural networks for fake news detection." arXiv preprint arXiv:1806.00749 (2018).

[12]  Bondielli, Alessandro, and Francesco Marcelloni. "A survey on fake news and rumour detection techniques." Information Sciences 497 (2019): 38-55.

[13]  Cardoso Durier da Silva, Fernando, Rafael Vieira, and Ana Cristina Garcia. "Can machines learn to detect fake news? a survey focused on social media." In Proceedings of the 52nd Hawaii International Conference on System Sciences. 2019.

[14]  Khattar, Dhruv, et al. "Mvae: Multimodal variational autoencoder for fake news detection." The World Wide Web Conference. 2019.

[15] Sharma, Karishma, Feng Qian, He Jiang, Natali Ruchansky, Ming Zhang, and Yan Liu. "Combating fake news: A survey on identification and mitigation techniques." ACM Transactions on Intelligent Systems and Technology (TIST) 10, no. 3 (2019): 1-42.

[16] Dong, Xishuang, Uboho Victor, and Lijun Qian. "Two-path Deep Semi-supervised Learning for Timely Fake News Detection." arXiv preprint arXiv:2002.00763 (2020).

[17] Goldani, Mohammad Hadi, Saeedeh Momtazi, and Reza Safabakhsh. "Detecting Fake News with Capsule Neural Networks." arXiv preprint arXiv:2002.01030 (2020).

[18] Kaliyar, Rohit Kumar, Anurag Goswami, Pratik Narang, and Soumendu Sinha. "FNDNet–A deep convolutional neural network for fake news detection." Cognitive Systems Research 61 (2020): 32-44.

[19] Yazdi, Kasra Majbouri, Adel Majbouri Yazdi, Saeid Khodayi, Jingyu Hou, Wanlei Zhou, and Saeed Saedy. "Improving Fake News Detection Using K-means and Support Vector Machine Approaches." International Journal of Electronics and Communication Engineering 14, no. 2 (2020): 38-42.

[20] Zhang, Jiawei, Bowen Dong, and S. Yu Philip. "Fakedetector: Effective fake news detection with deep diffusive neural network." 2020 IEEE 36th International Conference on Data Engineering (ICDE). IEEE, 2020.

[21] Zhou, Xinyi, Jindi Wu, and Reza Zafarani. "SAFE: Similarity-aware multi-modal fake news detection." arXiv preprint arXiv:2003.04981 (2020).